%% file: main.tex
\documentclass[conference]{IEEEtran}
\usepackage{times}

\usepackage[table]{xcolor}
\definecolor{lightcyan}{rgb}{0.88,1,1}

\usepackage[numbers]{natbib}
\usepackage{multicol}
\usepackage[bookmarks=true]{hyperref}
\usepackage[
    colorinlistoftodos,
    prependcaption,
    textsize=tiny,
    obeyFinal,
]{todonotes}
\usepackage{float}
\usepackage{graphicx}
\usepackage[label font=bf,labelformat=simple]{subfig}
\usepackage{booktabs}
\usepackage{xcolor}
\usepackage{amssymb}
\usepackage{lipsum}
\usepackage{breqn}

\IEEEoverridecommandlockouts

\begin{document}


\title{Which objects help me to act effectively? Reasoning about physically-grounded affordances}


\author{Anne Kemmeren*$^1$, Gertjan Burghouts*$^1$, Michael van Bekkum$^1$, Wouter Meijer$^1$, Jelle van Mil$^1$ 
\thanks{*These two authors contributed equally}
\thanks{
$^{1}$The authors are with TNO,
The Hague, Oude Waalsdorperweg 63, 2597 AK, Netherlands, {\tt \small\{anne.kemmeren, gertjan.burghouts, michael.vanbekkum, wouter.meijer, jelle.vanmil\}@tno.nl}}%
    } 
\maketitle

\input{content/0_abstract}
\IEEEpeerreviewmaketitle

\input{content/1_introduction}

\input{content/2_related_work}
\input{content/3_approach}
\input{content/4_results}

\input{content/5_conclusion}


\bibliographystyle{plainnat}
\bibliography{references}

\end{document}

%% file: content/0_abstract.tex
\begin{abstract}
For effective interactions with the open world, robots should understand how interactions with known and novel objects help them towards their goal. 
A key aspect of this understanding lies in detecting an object's affordances, which represent the potential effects that can be achieved by manipulating the object in various ways. 
Our approach leverages a dialogue of large language models (LLMs) and vision-language models (VLMs) to achieve open-world affordance detection. Given open-vocabulary descriptions of intended actions and effects, the useful objects in the environment are found. 
By grounding our system in the physical world, we account for the robot's embodiment and the intrinsic properties of the objects it encounters. In our experiments, we have shown that our method produces tailored outputs based on different embodiments or intended effects. The method was able to select a useful object from a set of distractors. Finetuning the VLM for physical properties improved overall performance. These results underline the importance of grounding the affordance search in the physical world, by taking into account robot embodiment and the physical properties of objects.

\end{abstract}

%% file: content/1_introduction.tex
\section{Introduction}
\label{sec:introduction}

To enable an intelligent robot to operate in the open-world, it needs to reason about how interacting with objects in the environment could contribute to its goal \cite{Gibson1977}. 
A crucial aspect of this capability is the robot’s awareness of object affordances: understanding which actions can be executed on an object and what effects those actions produce. The encountered objects could be both novel or well-known. 

Various previous works created models that endow robots with these reasoning skills by training on affordance datasets, where (regions of) objects are annotated with the actions that it allows \cite{luo2023grounded, zhang2024selfexplainable, zhai2021oneshot}. However, the datasets are annotated from a human point-of-view, and these approaches thus fail to address how the embodiment of the robot affects affordances. A door handle can only be turned if the robot has the correct type of manipulator. Moreover, these datasets solely consider object-action pairs and do not take the effects into account. For example, in Padv2 both a surfboard and a bed have the affordance to \textit{lie on} \cite{zhai2021oneshot}, but if the intended effect is to have the robot float on water only the surfboard is useful. Therefore, we believe that inclusion of the intended effect provides relevant task context, that is vital to distinguish which objects afford useful actions to the robot.

Embodied AI addresses both issues since it takes task context and robot embodiment into account. State-of-the-art models such as DreamerV3 \cite{hafner2023dreamerv3}, Octopus \cite{yang2023octopus} and SayCan \cite{saycan2022arxiv} have shown impressive performance to resolve what actions a robot should take to complete a (potentially complex) task, including situations where it needs to find and manipulate objects in the environment. However, these models consider only a very limited number of actions or skills that the robots can execute to keep the planning tractable. 
For example, DreamerV3, Octopus and SayCan would not be able to rotate the cap of a water bottle to retrieve water, since this specific grasp-and-rotation action is not in the list of possible actions. 

Our approach to open-vocabulary affordance detection considers a much wider range of actions. Yet, it keeps planning tractable by only returning an object that affords the action if it (1) contributes to the intended effect, and (2) the object can be manipulated given the robot embodiment. 
Following the work on Socratic Models \cite{chang2023prompting}, we propose a dialogue between a Large Language Model (LLM) and Vision Language Model (VLM). Given an open-vocabulary action and task, the dialogue finds the objects in the given image that can help the robot reach its goal. When reasoning about relevant objects, we were inspired by \cite{Tang2023CoTDetAK} to have the dialogue explicitly take physical properties of the object into account.  The novelty is that we prompt the LLM which objects provide the required affordance, while taking into account the limitations posed by the robot embodiment and the physical properties of object candidates as found by the VLM. 

\begin{figure}[t!]
    \centering
    \includegraphics[width=\columnwidth]{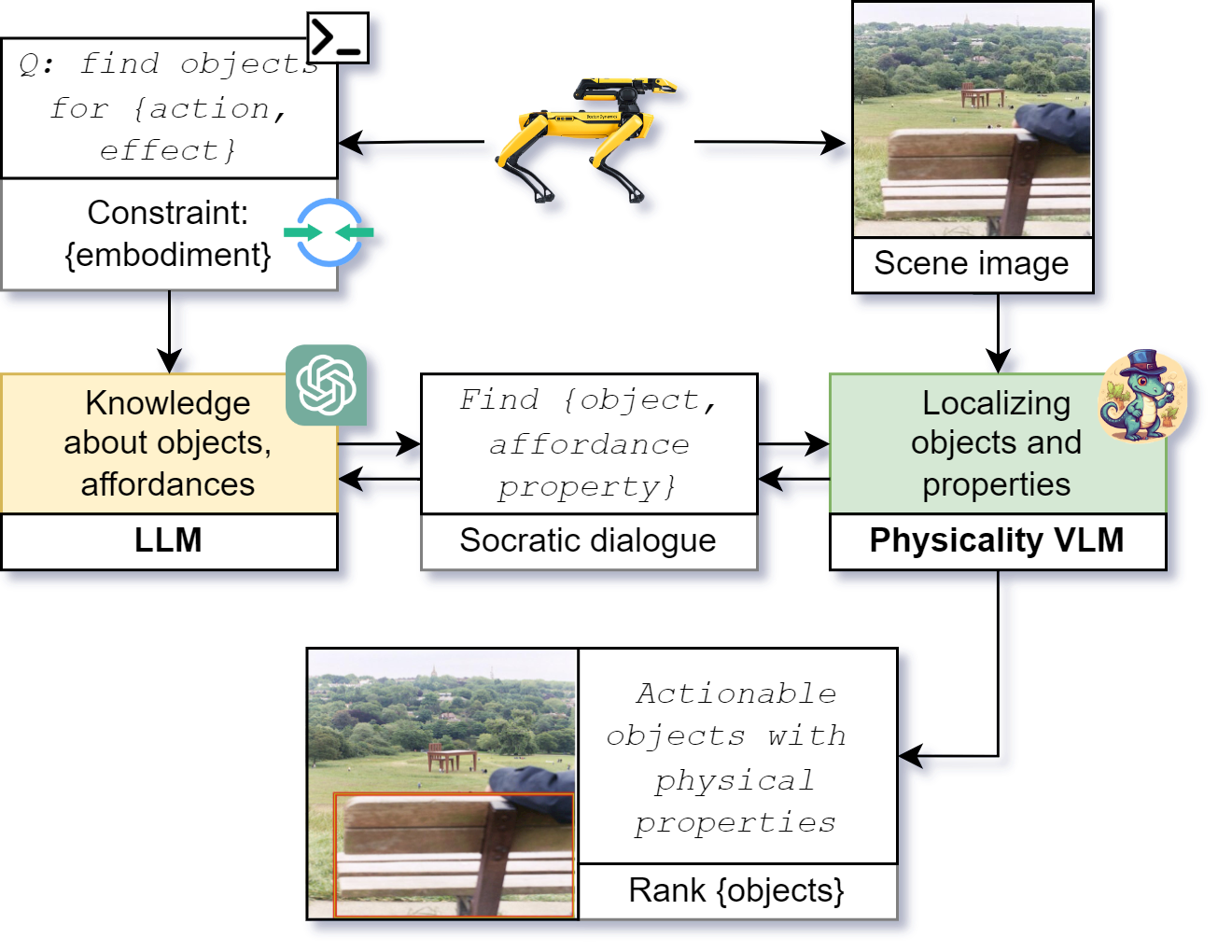}
    \caption{A socratic dialogue enriched with a physicality-grounded VLM can reason about relevant objects for the given task and action.}
    \label{fig:overview}
\end{figure}

\begin{figure*}
    \centerline{\includegraphics[width=\textwidth]{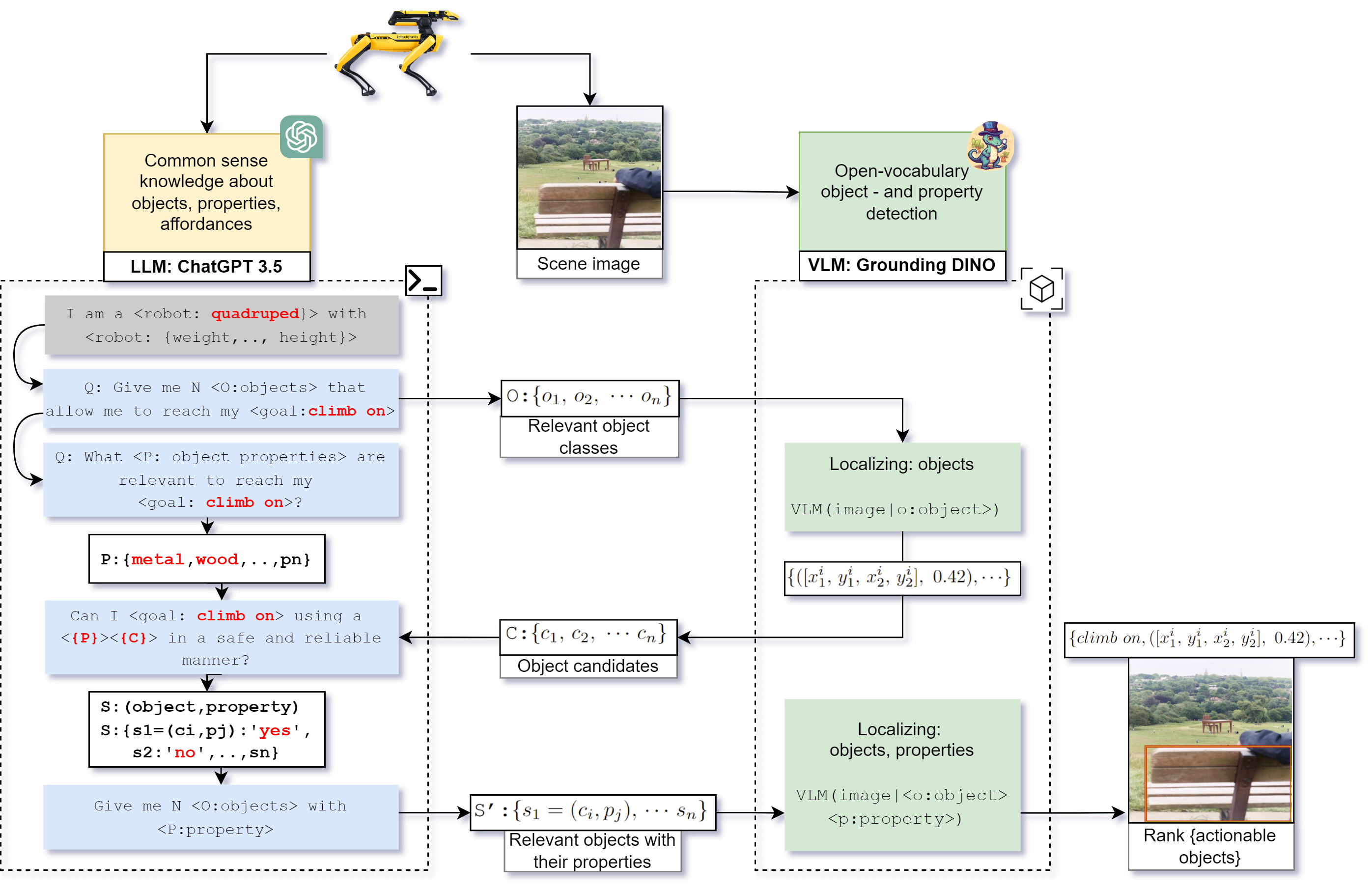}}
    \caption{The dialogue between an LLM (left) and VLM (right) reasons about what object in the given scene would give the quadruped robot the ability to climb to a better viewpoint.}
    \label{fig:dialog-details}
\end{figure*}

The contributions of this paper are as follows. (1) We develop a method for open-world affordance detection, where action, object and effect are all open-vocabulary. 
(2) We leverage out-of-the-box foundation models to reason about how embodiment of the robot and physical properties of the objects affect affordances.
(3) We validate the efficacy for physical properties, show the limitations, and how finetuning can improve this.
(4) In the experiments we show that the dialogue is able to more accurately select useful items from a set of distractors when the LLM takes physical properties of the object into account, and the VLM is finetuned on detecting objects with these properties, in an experiment with a real-life robot.  

%% file: content/2_related_work.tex
\section{Related Work} 
\label{sec:relatedwork}

Foundation models provide a generalized ability to reason about the physical world using everyday knowledge, understanding the behaviour and properties of objects (physical commonsense reasoning, \cite{ding2021dynamic}). However, physical grounding remains a challenge, while it is essential: a robot should know about the physical conditions of the objects of interest. For instance, the weight of the objects determines which ones can be picked up, whereas the material determines how to manipulate them (e.g. amount of exerted force). For this purpose, PG-VLM \cite{gao2023physically} improved the prediction of physical properties of BLIP-2 \cite{blip2}. BLIP2 is a foundation VLM, trained on a huge set of text-image pairs. However, it was shown that it could not estimate physical properties yet; the current very broad pretraining does not provide specialized knowledge about the object. PG-VLM incorporates such knowledge by instruction-tuning on physically grounded annotations. 

LLMs specifically have previously displayed semantic reasoning capabilities \cite{tang2023large}. Affordance learning  usually takes place via detecting visual representations or imitation learning \cite{bahl2023affordances}. Most robotic foundation models are however tailored to a specific embodiment \cite{firoozi2023foundation} and focus on movement or motion physics of objects \cite{Sun2023}. 

Dialogue models or Socratic Models are employed to support exploratory thinking through questioning \cite{chang2023prompting}. They allow back and forth interaction between foundation models as a dialogue in order to retrieve open-vocabulary affordances without fine-tuning 
\cite{zeng2022socratic}. 

The approach outlined in this paper also bolsters recent advancements in vision-language-action models such as RT-2 \cite{rt22023arxiv} and RT-2-X \cite{embodimentcollaboration2024open}. Where these developments enable robots to execute open-vocabulary actions, our work addresses where in the environment such open-vocabulary actions could be executed by leveraging open-vocabulary affordance detection.

%

%% file: content/3_approach.tex
\section{Approach} 
\label{sec:approach}

\subsection{Problem Definition}
Given a particular goal and environment, the robot should be capable of interacting with objects in the environment, so that it reaches the goal. Therefore, we consider the problem of detecting the objects in the environment that afford the robot to take relevant actions towards its goal. As we operate in the open world, we require action and goal descriptions to be open-vocabulary.

\subsection{Dialogue Overview}
By chaining LLMs and VLMs, the specification of objects, actions and goals can all be done in free-form text. Figure \ref{fig:dialog-details} provides a detailed overview of the dialogue implementation. The LLM is prompted to acquire a set of objects that allow for the action or goal, while taking the constraints into account that are posed by the robot embodiment, mission requirements (e.g. safety) and properties of the objects (e.g. material). The VLM is queried to find the objects that have the right properties in the images captured by the robot's camera. The found objects are checked by the LLM regarding which properties they should have in order to be successful. 

\subsection{Dialogue Configuration}
The LLM is configured with two variables to ground its reasoning in the real world. The first variable is information about robot embodiment and the second one is which physical object properties the LLM should consider to assess if an object will be relevant to the completion of the goal. 

\textit{Robot}: To take the robot embodiment into account,  we provide a textual description of the robot. This allows the LLM to reason about the robot's capabilities and limitations when suggesting objects to interact with. This description is based on the specifications of the robot platform and includes properties such as robot type (e.g. wheeled, legged), dimensions (e.g. height, width), type of manipulator and weight:

\begin{equation}
\begin{split}
    \texttt{Robot} = \{ & type \rightarrow quadruped,\\
    & weight \rightarrow 50 kg, \cdots,\\ 
    & height \rightarrow 50 cm\}
\end{split}
\end{equation}

\textit{Objective}: The goal and requirements are specified:

\begin{equation}
\begin{split}
    \texttt{Requirements} = \{ & goal \rightarrow climb\,\,on,\\
    & conditions \rightarrow \{safe,\,\,reliable\}\,\}
\end{split}
\end{equation}

Note that the conditions can be defined in a soft manner, because the LLM can deal with such descriptions.

\textit{Objects}: The physical properties of an object determine for a large part which interactions are allowed, e.g.: a robot might be able to stand on a metal box, but not on a paper box. We specifically ask the LLM to consider what object properties are relevant for the given action and goal, such that the VLMs can subsequently find only those object instances that have these properties. We provide a list of the properties and values that the VLMs on the robot could potentially detect, such as material types and colors.

\begin{equation}
\begin{split}
    \texttt{Properties} = \{ & colors \rightarrow \{blue, \cdots, green\}, \cdots\\
    & material \rightarrow \{plastic, \cdots, metal\}\,\}
\end{split}
\end{equation}

\begin{table*}
\centering
\caption{\textbf{Intended effect}: For the same Action, but different intended Effects, our method suggests different objects.}
\label{table:reasoning}
\begin{tabular*}{\linewidth}{@{\extracolsep{\fill}} l c c c c}
\toprule  
    \textit{Action} & \multicolumn{2}{l}{Contain to get:} & \multicolumn{2}{l}{Stand on to:} \\
    \textit{Effect} & liquids from A to B & groceries from A to B & increase robot's height & float on water \\
    \midrule

    Bowl        & \checkmark    & \checkmark    & -             & - \\
    Box, Bucket & \checkmark    & \checkmark    & \checkmark    & - \\
    Blender, Can, Carton, Cup 
                & \checkmark    & -             & -             & - \\
    Jar, Kettle, Mug, Tray, Vase 
                & \checkmark    & -             & -             & - \\
    Bag, Belt   & -             & \checkmark    & -             & - \\
    Bench       & -             & \checkmark    & \checkmark    & - \\
    Bottle, Ladder, Stool, Book
                & -             & -             & \checkmark    & - \\
    Basket      & -             & \checkmark    & \checkmark    & \checkmark \\
    
   \bottomrule
\end{tabular*}
\vspace{-2mm}
\end{table*}

\begin{table*}
\centering
\caption{\textbf{Embodiment}; For a different embodiment or action, our method suggests other object properties.}
\label{table:reasoning2}
\begin{tabular*}{\linewidth}{@{\extracolsep{\fill}} l l l l }
\toprule  
     & Small robot stands on & Large robot stands on  & Large robot places a small object on  \\
    \midrule

    Basket & \{plastic, metal\}        & \{plastic, metal\}        & \{plastic, metal\} \\
    Bench  & \{plastic, metal\}        & \{plastic, metal\}        & \{plastic, metal, wood, glass\} \\
    Box     & \{plastic\}               & \{plastic\}               & \{plastic, metal, wood, paper\} \\
    Book    & \{plastic, paper\}        & \{\}                      & \{\} \\
    Ladder  & \{plastic, metal\}        & \{plastic, metal\}        & \{\} \\
    Stool   & \{plastic, metal, wood\}  & \{plastic, metal\}        & \{plastic, metal, wood, paper\} \\

   \bottomrule
\end{tabular*}
\vspace{-2mm}
\end{table*}

\begin{table*}
\centering
\caption{\textbf{Adaptation}: Adapting a VLM to the properties of objects is effective, increasing mAP. }
\label{table:object_detection}
\begin{tabular*}{\linewidth}{@{\extracolsep{\fill}} l l l l l l l l l l l l l l l l }
\toprule  
     & \multicolumn{4}{l}{Wood} & 
     \multicolumn{1}{l}{Paper} & 
     \multicolumn{2}{l}{Plastic} & 
     \multicolumn{4}{l}{Metal} & \\
     VLM & 
     {\footnotesize Basket} & {\footnotesize Stool} & {\footnotesize Ladder} & {\footnotesize Bench} & 
     {\footnotesize Box} & 
     {\footnotesize Stool} & {\footnotesize Basket} & 
     {\footnotesize Ladder} & {\footnotesize Basket} & {\footnotesize Stool} & {\footnotesize Bench} & Avg. \\
    \midrule
    As-is & 0.12 & 0.44 & 0.32 & 0.56 & 0.07 & 0.10 & 0.13 & 0.21 & 0.28 & 0.43 & 0.28 & 0.27  \\
    \rowcolor{lightcyan} Physicality & 0.53 & 0.66 & 0.37 & 0.66 & 0.46 & 0.13 & 0.25 & 0.34 & 0.32 & 0.43 & 0.48 & \textbf{0.42}  \\
   \bottomrule
\end{tabular*}
\vspace{-2mm}
\end{table*}

\begin{table*}
\centering
\caption{\textbf{Generalization}: The adapted VLM improves the prediction (measured by mAP) of properties of unseen objects (mAP). }
\label{table:object_detection_unseen}
\begin{tabular*}{\linewidth}{@{\extracolsep{\fill}} l l l l l l l l l l l l l }
\toprule  
     & 
     \multicolumn{3}{l}{Plastic} & 
     \multicolumn{2}{l}{Glass} & 
     \multicolumn{2}{l}{Wood} & 
     \multicolumn{4}{l}{Metal} & \\
     VLM & 
     {\footnotesize Lamp} & {\footnotesize Crate} & {\footnotesize Hammer} & 
     {\footnotesize Lamp} & {\footnotesize Laptop} & 
     {\footnotesize Lamp} & {\footnotesize Crate} & 
     {\footnotesize Lamp} & {\footnotesize Crate} & {\footnotesize Laptop} & {\footnotesize Laptop} & 
     Avg. \\
    \midrule
    As-is & 0.45 & 0.05 & 0.15 & 0.12 & 0.29 & 0.11 & 0.02 & 0.42 & 0.11 & 0.17 & 0.43 & 0.21 \\
    \rowcolor{lightcyan} Physicality & 0.73 & 0.14 & 0.18 & 0.13 & 0.38 & 0.12 & 0.11 & 0.51 & 0.19 & 0.51 & 0.65 & \textbf{0.33} \\
   \bottomrule
\end{tabular*}
\vspace{-2mm}
\end{table*}

\subsection{Reasoning}

During runtime, the dialogue has as input a set of images that were collected of the environment, a (sub)goal specification and a desired action. Its output is a set of predictions of the objects that afford the desired action and contribute to reaching the goal.

The LLM is used in chain mode, such that earlier prompts and responses are stored as context. Our dialogue starts by informing the LLM of the context from the previous subsection:

\begin{equation}\label{eq:robot}
\begin{split}
    \texttt{I am a <}robot : type\}\texttt{> with}\\\texttt{<}robot:\{weight, \cdots, height\}\texttt{>}
\end{split}
\end{equation}

The LLM is prompted with the question which $N$ objects can reach the goal. Its response is a text that includes a list of objects. The text is parsed to extract the names of the objects that are potentially suitable. The VLM is tasked to find these object names by prompting it with these names as labels. The threshold is set low (0.3) to avoid false negatives. For all object candidates provided by the VLM, the LLM is prompted whether the specific object can solve the task. 

The LLM is queried for the object properties that are relevant to solve the task. For instance, object color is not relevant to climb on the object, but its material is. For the set of properties that is considered relevant, the specific instances are retrieved. E.g., object materials can be metal, wood, plastic, etc. The LLM is prompted which combinations of the object class with the relevant properties can solve the task. For instance, a prompt `can the robot stand on a metal box in a safe and reliable manner?' More formally:

\begin{equation}\label{eq:task}
\begin{split}
    \texttt{Can I <}goal\texttt{> using a <}property\texttt{> <}object\texttt{>}\\\texttt{in a <}conditions\texttt{> manner?}
\end{split}
\end{equation}

The response is parsed by extracting affirmative or negative words to understand if the object-property combination is suitable for the task at hand. The VLM is prompted for the set of suitable object-property combinations. 

\begin{equation}\label{eq:vlm}
\begin{split}
    \texttt{VLM(image}\,\,|\,\,\texttt{<}property\texttt{>}\,\,\texttt{<}object\texttt{>)} \rightarrow \\\{([x_1^i,\,y_1^i,\,x_2^i,\,y_2^i],\,\,c), \cdots \}
\end{split}
\end{equation}

Here are $x_1^i$, $y_1^i$ the x, y of the upper-left position in the image, $x_2^i$, $y_2^i$ the x, y of the lower-right position in the image, and $c$ the confidence value. These predictions are the output of our dialogue.






%% file: content/4_results.tex
\section{Experiments} 
\label{sec:experiments}

We analyze the capabilities of our method, by answering the following questions: 

\begin{enumerate}
    \item For a given action, will it search for different objects when the intended effect is different?
    \item For another embodiment of the robot, will it search for different objects with other properties?
    \item Can a VLM designed for object detection be finetuned to estimate object properties? Does that generalize to unseen objects?
    \item How well does our method find the right objects in the wild?
\end{enumerate}

\subsection{Setup}

The LLM is ChatGPT 3.5 \cite{chatgpt}. The VLM is Grounding DINO \cite{liu2023grounding}, because it can detect objects (i.e. localization). PG-BLIP  \cite{gao2023physically} is also interesting, but it can only classify images (i.e. no localization) and it is a very large model which is disadvantage for deployment on a robot. For evaluation, we consider the PACO image dataset \cite{ramanathan2023paco}, because it has annotations of the objects, parts and their properties including materials. We also include an experiment with the SPOT robot in our Open-World Robotics lab.

\subsection{Effect-specific Objects}

For a given action, but for a different intended effect, our method suggests different objects. Table \ref{table:reasoning} shows the results after evaluating Equations \ref{eq:robot} and \ref{eq:task} for the respective Actions and Effects in the table header (in a combined prompt of Action + Effect in Equation \ref{eq:task}). When the desired effect is to get liquids from one location to another location, the intended action is contain, and suitable objects are a Bowl, Can or Vase. However, if the action is the same, but the intended effect is to bring groceries to another location, a Can is not suitable, while a Bag or Basket are suitable. Likewise, for a robot that is tasked to stand on something, the intended effect matters. There is a difference when the intended effect is to increase the robot's height (e.g. Bucket, Bench, Basket), or that the robot should float on water (only Basket). Our method can handle various intended effects.

\subsection{Constraints of the Embodiment}

For a robot with a different embodiment, our method yields objects with different properties. Table \ref{table:reasoning2} shows the objects and their properties as suggested by our method, for respectively a small robot (height of 25 centimeters and 5 kilograms) and a large robot (height of 50 centimeters and 50 kilograms). This is to evaluate the effect of Equation \ref{eq:robot} on Equation \ref{eq:task}. The method indicates that the two robots can stand on different objects with different properties. The small robot can stand on a Book, whereas the large robot cannot. The small robot can stand on a Wooden Stool, whereas the large robot can only stand on Metal and Plastic Stools. The intended action also matters: a small object can be placed on a Glass Bench and Paper Box, while both robots cannot stand on these objects.

\begin{figure*}[h!]
    \begin{minipage}{\textwidth}
        \includegraphics[width=\textwidth]{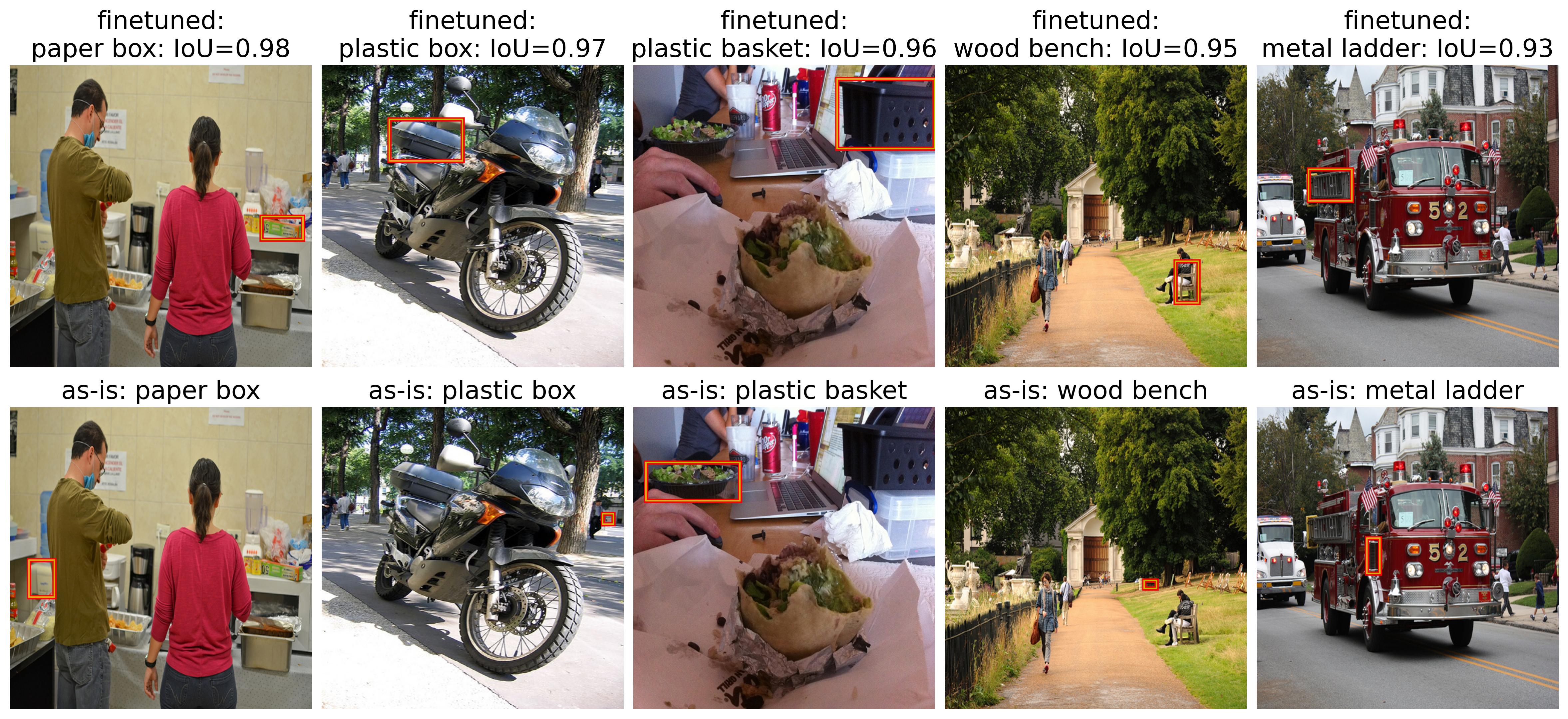}
        \caption{\textbf{Improvements}: After adaptation, the VLM's prediction of objects and properties is improved (top row).}
        \label{fig:improvements}
    \end{minipage}
    \begin{minipage}{\textwidth}
        \includegraphics[width=\textwidth]{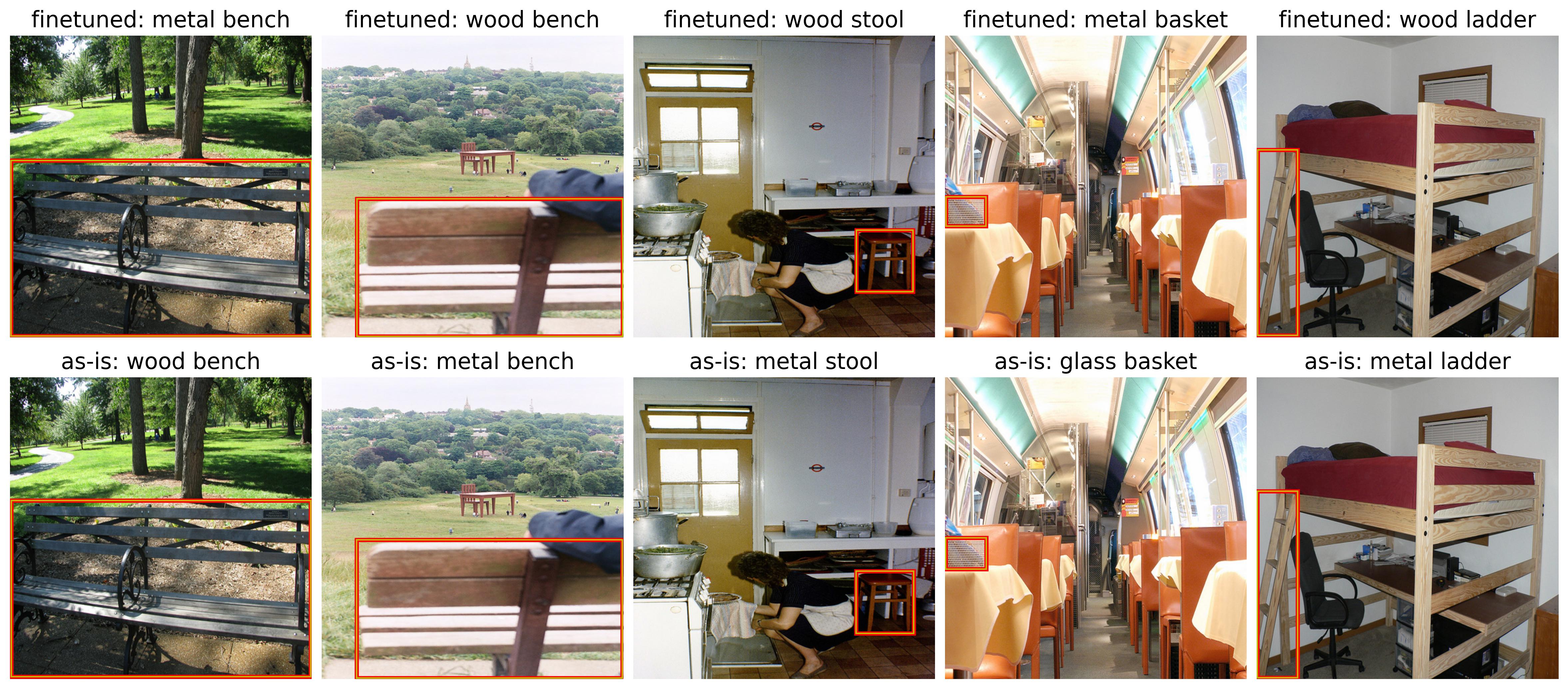}
        \caption{\textbf{Object properties}: Correcting the wrong properties for several object classes.}
        \label{fig:attributes}
    \end{minipage} 
    \begin{minipage}{\textwidth}
        \includegraphics[width=\textwidth]{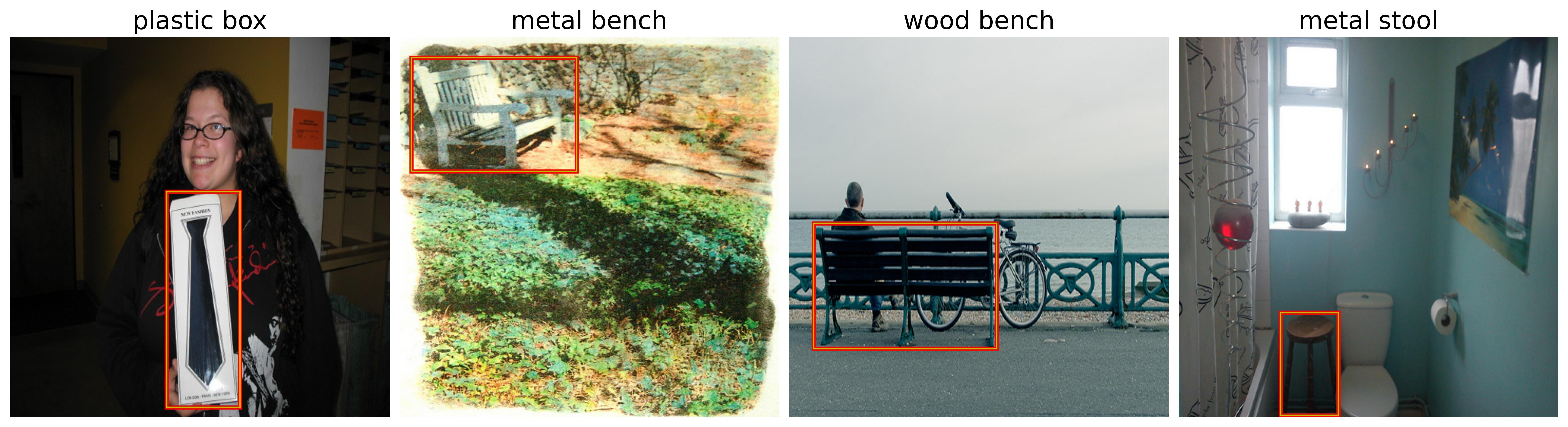}
        \caption{\textbf{Errors}: Remaining errors such as a Metal Bench was is confused with a Wooden Bench (third image) and a Metal Stool (right) which is a Wooden Stool but it has Metal legs.}
        \label{fig:errors}
    \end{minipage}
\end{figure*}

\begin{figure*}[t!]
    \centering
    \subfloat{\includegraphics[height=4.4cm]{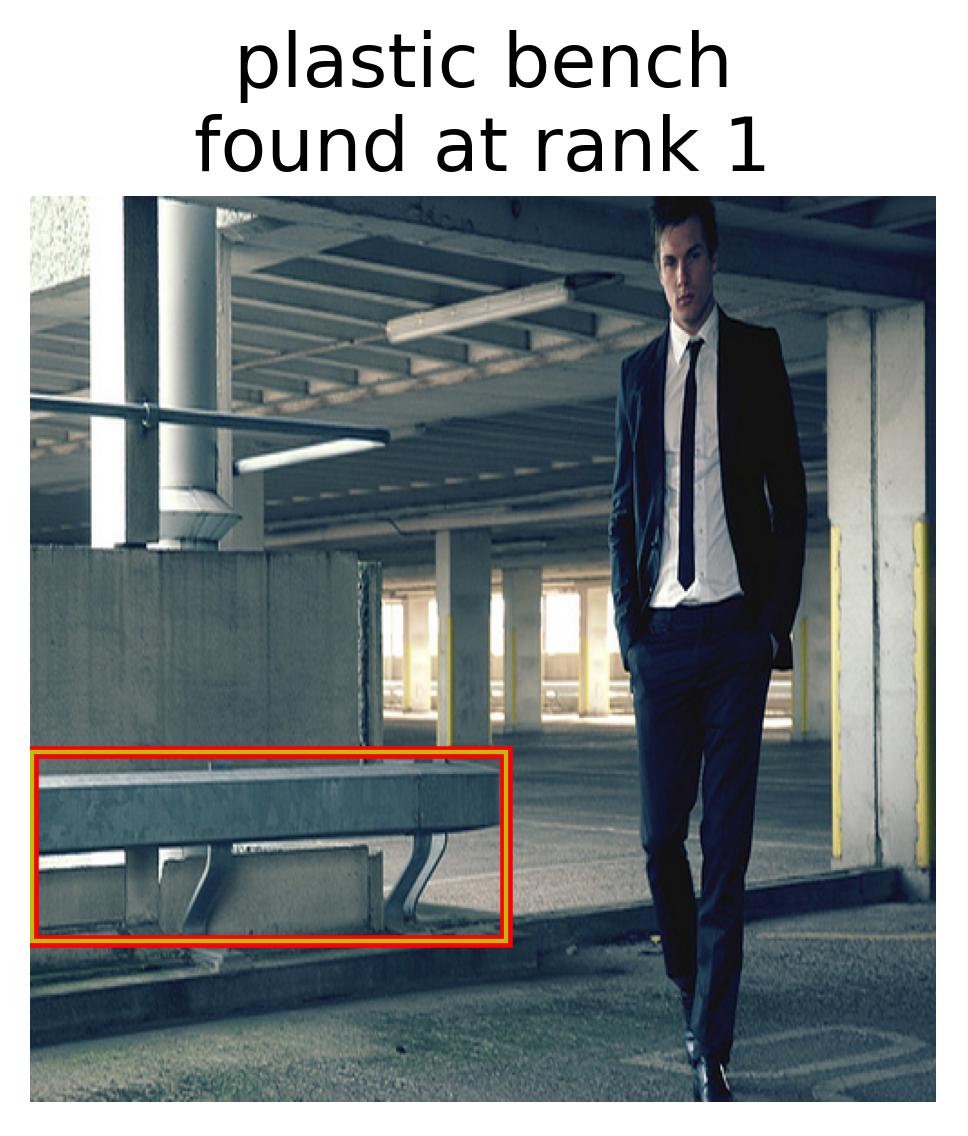}} \hspace{1cm}
    \subfloat{\includegraphics[height=4cm]{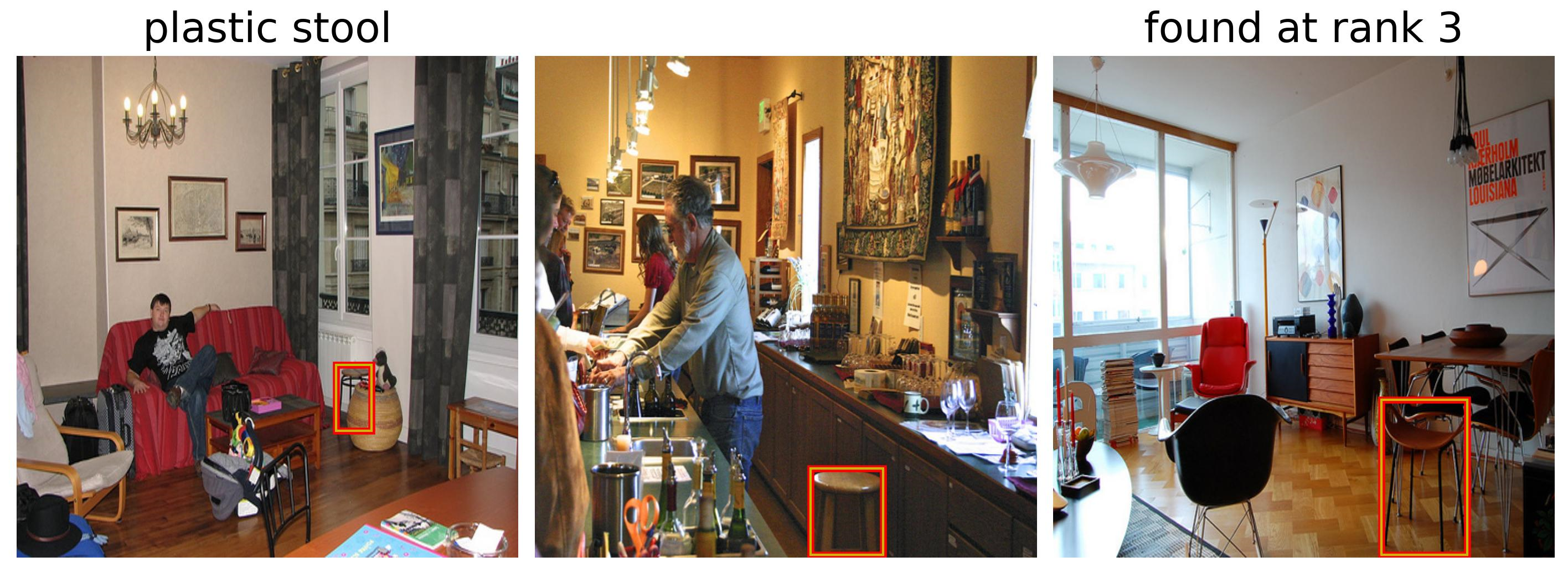}}
    \caption{\textbf{Sorting}: A Plastic Bench is found directly i.e. rank 1 (left), whereas the Plastic Stool is found at rank 3 (right) after finding Stools with the wrong properties.}
    \label{fig:ranks}
\end{figure*}

\begin{figure}[t!]
    \centerline{\includegraphics[width=\linewidth]{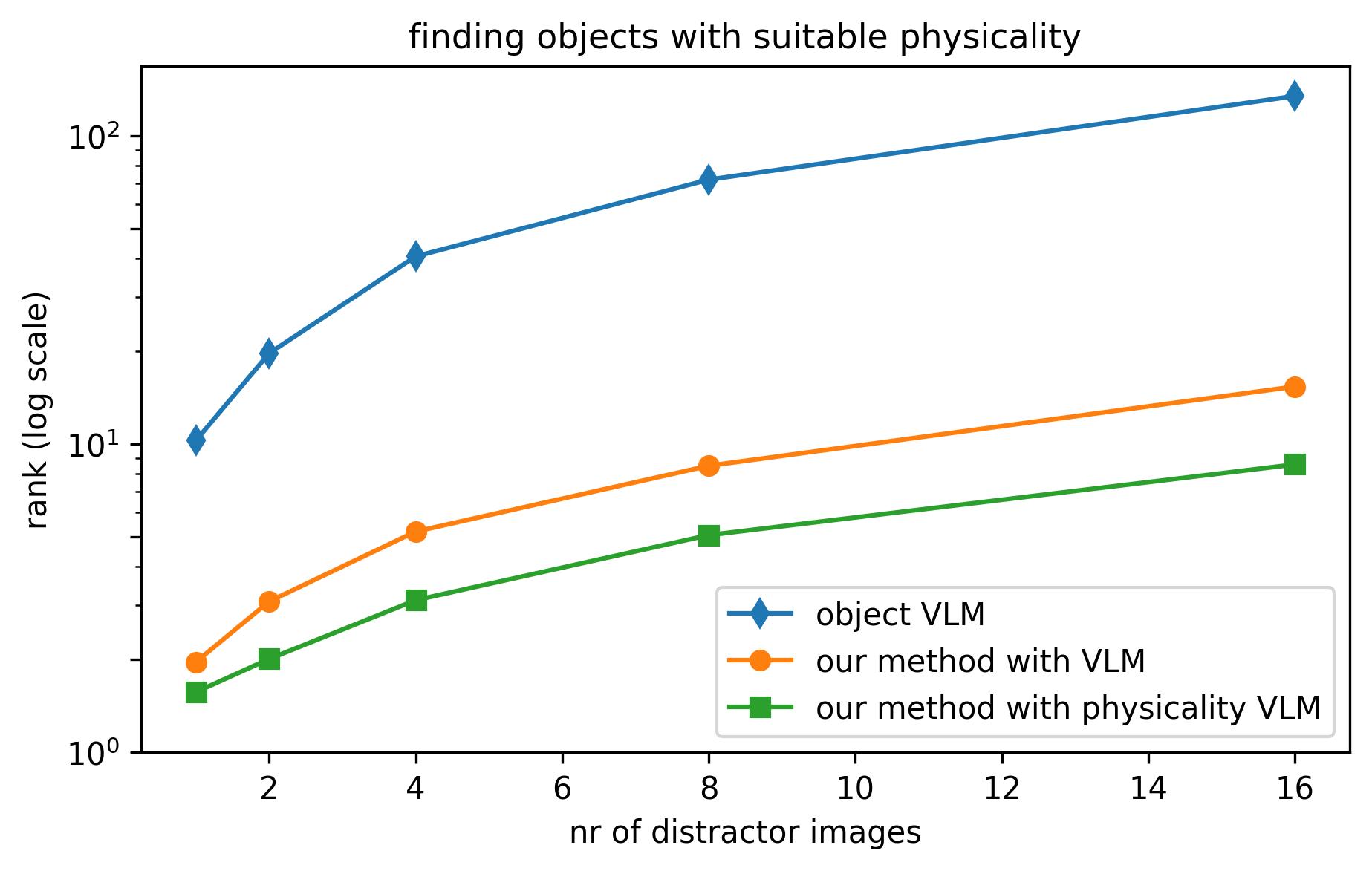}}
    \caption{\textbf{Effectiveness}: Our method with adapted VLM finds the right objects with suitable properties much faster than the object detection VLM and our method without adapted VLM.}
    \label{fig:finding}
\end{figure}

\subsection{Improving Detection of Object Properties}

Given the importance of objects and their properties: how well can a VLM detect objects with specific properties? Table \ref{table:object_detection} shows the results for our VLM on the first row. The performance is not good: mAP=0.27. Especially the Paper Box (mAP=0.07), Wood Basket (mAP=0.12) and Plastic Stool (mAP=0.10) are hard to detect. 

Finetuning is applied with the goal to improve the performance of estimating object properties. The subset of PACO as shown in Table \ref{table:object_detection} is leveraged for this purpose. We follow Grounding DINO's standard recipe for finetuning, i.e. 15 epochs with the provided learning rate and schedule. With finetuning, the results can be improved significantly, from mAP=0.27 to mAP=0.42 on average. Paper Box is improved from mAP=0.07 to mAP=0.46, whereas Plastic Stool is not improved much: mAP=0.10 to mAP=0.13. Plastic Stool is a rare object-material combination, which makes it hard to learn. Wood Basket is improved from mAP=0.12 to mAP=0.53.

\subsection{Generalization to Unseen Object-Properties}

The question is whether the newly learned object properties generalize to unseen objects. There is a performance gain for the unseen objects from Table \ref{table:object_detection_unseen}. On average, the VLM performance of mAP=0.21 is increased to mAP=0.33. Some objects do not improve much, e.g. Plastic and Metal Crate; Wood and Glass Lamp. These are hard objects because they are respectively partially visible (crates often are in between other objects) and small. Paper Box, Wood Stool and Metal Ladder are improved by large margins, without having seen these object-material combinations during training. 

\subsection{Analysis of Results}

We inspect the objects and properties for which the improvement is most significant. Figure \ref{fig:improvements} shows the largest gains, sorted from most to less gain. In all cases, the VLM as-is predicts the objects wrongly, mAP=0. The finetuned VLM with physical properties predicts the objects and their properties well, even in very challenging circumstances, e.g. the Metal Ladder on the back of the firetruck photographed under a shear viewing angle (Figure \ref{fig:improvements}, right). The Wood Bench and the Paper Box (left) are small, whereas the Plastic Box on the motorcycle is in the midst of clutter. 

Most object predictions have the correct object label, but a wrong material label. This is to be expected: it is easier to assess the object class than its properties. Also, the pretraining of most VLMs is focused on object classes, not on object properties. We inspect which object properties have improved most, see Figure \ref{fig:attributes}. The Wood Bench is corrected to a Metal Bench (left) and vice versa (second column). Glass Basket is corrected to Metal Basket (fourth column) and Metal Ladder to Wood Ladder (right).

Figure \ref{fig:errors} shows several examples of remaining errors. The Plastic Box (left) is probably correct; it appears to be a mislabeling in the groundtruth. The Wood Bench is mistaken for a Metal Bench (second image) and vice versa (third image). The prediction Wood Bench (third image) is actually a Metal Bench, but it is hard to distinguish, even for humans. The Metal Stool (fourth column) is a Wood Stool but it has metal legs.

\begin{figure*}[t!]
    \centerline{\includegraphics[width=\textwidth]{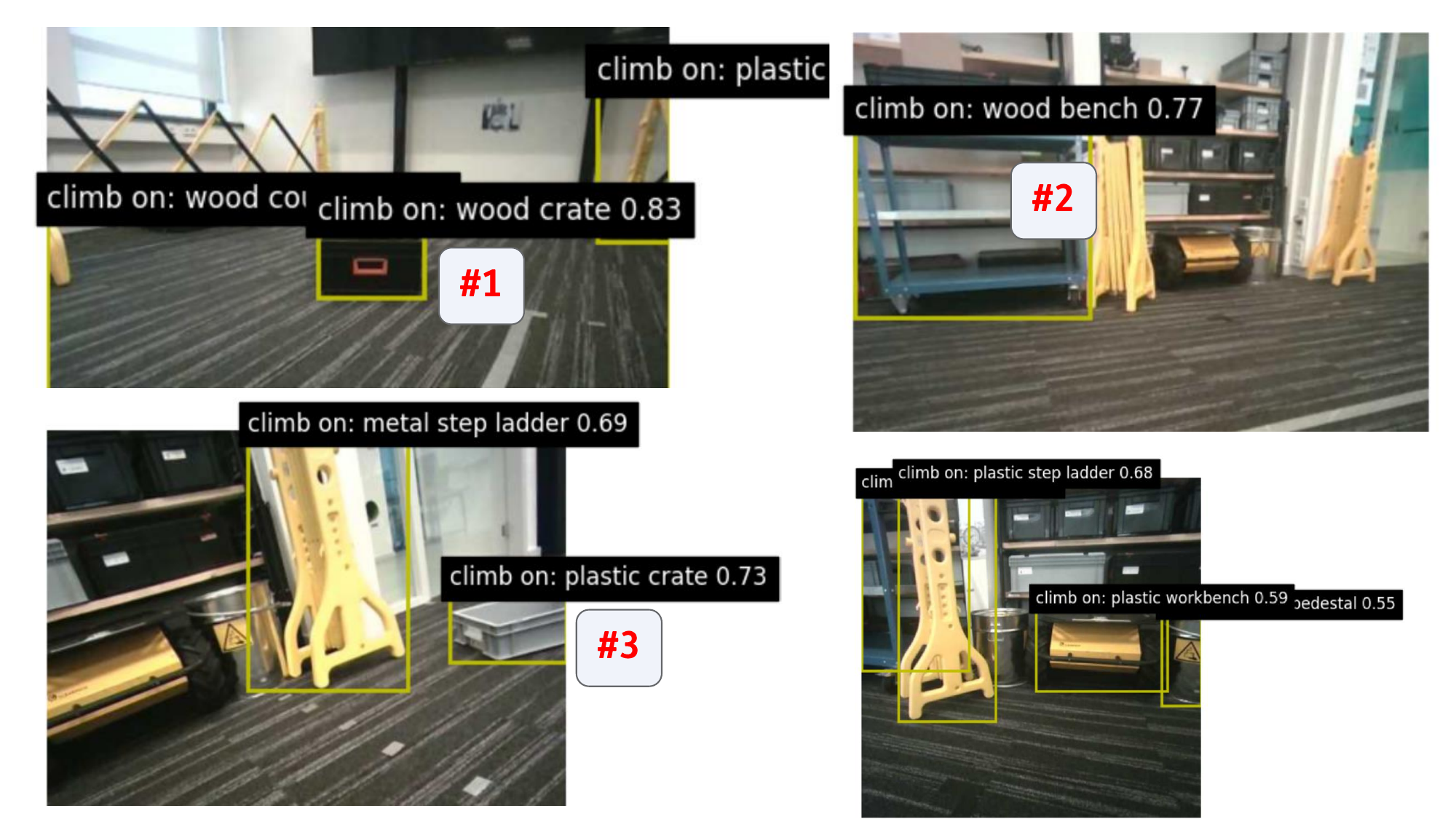}}
    \caption{\textbf{Robot}: With our method, the robot finds a Wooden Crate (top left) when the intended action is `climb on'.}
    \label{fig:robotlab}
\end{figure*}

\subsection{Searching for the Right Objects}

Our method is a dialogue between an LLM and a VLM to find the right objects with the desired properties to fulfil an action for an intended effect. We evaluate how well our method can find the desired combination of the object and the property. This is to validate the joint efficacy of Equations \ref{eq:robot}, \ref{eq:task} and \ref{eq:vlm}. 

If the desired object is Paper Box, we collect distractor images from PACO that contain Boxes with other properties than Paper, e.g. a Plastic Box. The objective is to find the Paper Box in the target image, in the midst of $N$ images that contain other Boxes. We increase $N$ progressively to assess how well our method can find the desired object when the task becomes increasingly difficult. This is to mimic that the robot's environment increases. We repeat this trial for all images and combinations from Table \ref{table:object_detection} and average the results. The results are shown in Figure \ref{fig:finding}, with the $N$ distractor images on the $x$-axis and the rank at which the desired object is found on the $y$-axis (log scale). The relation between the amount of distractors and the rank is approximately linear, as expected. The blue line (`object VLM') is the performance when only the object class is taken into account. Evidently, this is not an adequate strategy: the objects with the right properties are found at ranks $>$ 10, e.g. with 8 distractor images it is found around rank 70 on average. Our method takes the desired properties into account. Even with the VLM as-is (orange line), which is not optimal for predicting object properties, the ranking is significantly improved. With 8 distractor images, the rank improves from 70 to $<$9. It can be concluded that it is beneficial to take the object properties into account. When our method is combined with the finetuned VLM (green line), the average rank further decreases from 9 to 4.5. This means that the efficiency of finding the object with the right properties is further improved by a factor of 2 on average. 

To find an object with the right property, is illustrated in Figure \ref{fig:ranks}, for three examples (rows), respectively found at ranks 1 (optimal), 3 and 5. A Plastic Bench is found at rank 1 (top row). A Plastic Stool (middle row) is found at rank 3. At ranks 1 and 2, other Stools are found, which are not Plastic but Wood and Metal. A Wood Basket (bottom row) is found at rank 5. Again, the objects are all Baskets, but not the right property, e.g. a Plastic Basket. 


\subsection{Our Method on a Robot}

Our final experiment is to equip a robot with our method. The robot is Spot by Boston Dynamics. We task it to search for objects to climb on, e.g. to increase its height and look over obstacles. It is remotely controlled past 16 positions that are a few meters apart. At each position, an image is recorded by its omni-directional cameras. Some examples are provided in Figure \ref{fig:robotlab}. Our method is applied to the collected images. In the 16 images, there are several distractor objects such as cables, desks, equipment, tape, etc. 

Our method is searching for objects that can fulfil `climb on' (e.g. to look over some obstacle). Equations \ref{eq:robot} is initialized with the SPOT specifications, Equation \ref{eq:task} is invoked with `climb on' and `safe', with Grounding DINO for Equation \ref{eq:vlm}. For this task, our method has ranked the most suitable objects and their properties. At rank 1, it finds a Wood Crate (top left), although actually it is a composite material. At rank 2, it suggests a Wood Bench (top right); and at rank 3 (bottom left) it suggests a Plastic Crate. The results are not perfect, but the suggested objects are sensible and can serve the purpose. It shows the potential of our method in practice.

%% file: content/5_conclusion.tex
\section{Discussion and Conclusion} 
\label{sec:conclusion}
We proposed a dialogue of out-of-the-box foundational models (LLM and VLM) to find objects in the open world that afford desired actions that contribute to a specific robot's goal. We ground the responses of the LLM and VLM by taking the robot embodiment and physical object properties into account. 

The results show that the framework can successfully localize the relevant objects, while taking into account robot embodiment and goal context. By forcing the LLM and VLM to reason about and detect relevant object properties, the method finds objects that are more useful to the task than a naive approach without the dialogue. Detecting objects with relevant properties is further improved by specifically finetuning the VLM on e.g. materials. There is still a performance gap as the VLM still struggles with distinguishing between subtle object properties, especially for small objects or objects that are often (partially) obscured. The current finetuned VLM already improves the search for the right objects with suitable properties.

As future work, we consider a number of ways to improve the proposed dialogue.
The VLM currently finds objects that have a certain property, e.g. a stool that is made from wood. However, this does not consider that objects can be composed of different parts that have different properties. The work could therefore be extended to allow the VLM to find objects that have a mixture of properties (e.g. a stool made from wood and metal), or to query the LLM if an object property is relevant to some \textit{part} of the object.
Moreover, the dialogue has as input an open-vocabulary action, in combination with the intended effect. The dialogue can be edited to have the action-object tuple as output instead. Then, by only giving a textual specification of the goal, the framework could output open-vocabulary actions on relevant objects.  
Lastly, the LLM now manipulates class labels to find objects that are useful in the queried setting. Inspired by \cite{Tang2023CoTDetAK}, the method could become more nuanced if considering the attributes that make an object useful instead. Then, any object with a useful attribute can be found, hence reducing reliance on the LLM to generate the correct class labels. 
